# IEEE Copyright Notice

# A Unified Framework to Classify Business Activities into International Standard Industrial Classification through Large Language Models for Circular Economy

Xiang Li[1], Lan Zhao[1,2], Junhao Ren[1], Yajuan Sun[2,*], Chuan Fu Tan[2], Zhiquan Yeo[2], Gaoxi Xiao[1]
[1]School of Electrical and Electronic Engineering, Nanyang Technological University, Singapore
[2]Singapore Institute of Manufacturing Technology (SIMTech), Agency for Science, Technology and Research (A*STAR), Singapore
Emails: {xiang002, zhao0468, junhao002}@e.ntu.edu.sg, {Sun_Yajuan, Tan_Chuan_Fu, zqyeo}@simtech.a-star.edu.sg, egxxiao@ntu.edu.sg

*Abstract* - **Effective information gathering and knowledge codification are pivotal for developing recommendation systems that promote circular economy practices. One promising approach involves the creation of a centralized knowledge repository cataloguing historical waste-to-resource transactions, which subsequently enables the generation of recommendations based on past successes. However, a significant barrier to constructing such a knowledge repository lies in the absence of a universally standardized framework for representing business activities across disparate geographical regions. To address this challenge, this paper leverages Large Language Models (LLMs) to classify textual data describing economic activities into the International Standard Industrial Classification (ISIC), a globally recognized economic activity classification framework. This approach enables any economic activity descriptions provided by businesses worldwide to be categorized into the unified ISIC standard, facilitating the creation of a centralized knowledge repository. Our approach achieves a 95% accuracy rate on a 182-label test dataset with fine-tuned GPT-2 model. This research contributes to the global endeavour of fostering sustainable circular economy practices by providing a standardized foundation for knowledge codification and recommendation systems deployable across regions.**

*Keywords* – **Muti-class classification, Large Language Models (LLMs), International Standard Industrial Classification (ISIC)**

## I. INTRODUCTION

Waste recycling and reusing are promising practices to promote Circular Economy (CE). In industries, reusing waste or by-products from one company as a resource for another company is defined as Industrial Symbiosis (IS) [1]. It engages traditionally separate industries in a collective approach via physical exchanges of materials, water, and energy. Through waste-to-resource matching, amount of waste generated and need for virgin materials is reduced, creating both economic and environmental benefits. As the world grapples with the challenges of resource scarcity and environmental degradation, industrial symbiosis offers a path towards more sustainable and circular industrial practices.

Existing industrial symbiosis parks around the world exemplify the potential of this concept. These parks are dedicated hubs where industries come together to engage in resource sharing, such as Kalundborg Symbiosis [2] and Xiong'an New Area [3]. Each of these parks showcases the adaptability of industrial symbiosis to different economic, social, and environmental contexts. Developing a unified knowledge repository on existing cases not only helps analyse various aspects of industrial symbiosis, but also reveal new opportunities across disparate regions. However, a signification barrier to construct such a knowledge repository lies in the absence of a universally standardized framework for representing business activities as different countries have their own industrial classification standards. For example, Singapore utilizes the Singapore Standard Industrial Classification (SSIC) and Europe employs Nomenclature statistique des activités économiques dans la Communauté européenne (NACE) for economic activity categorization. This hinders broader adoption of industrial symbiosis and limits the opportunities for cross-border collaboration.

To address this challenge, this paper leverages Large Language Models (LLMs) to classify textual data describing economic activities into the International Standard Industrial Classification (ISIC) [4] framework. Such an approach enables any economic activity description provided by businesses worldwide to be categorized into the unified ISIC framework. We focus on ISIC due to the following reasons. Firstly, ISIC represents a globally recognized framework for the categorization of economic activities, anchored in globally accepted economic principles. Its primary objective is to present a transparent and inclusive categorization for activities, facilitating the collection and presentation of statistics in a format tailored for economic scrutiny, decision-making, and policy formulation [5], [6]. Secondly, a significant number of nations either adopt this classification framework directly or implement one influenced by it. Owing to its extensive adoption, ISIC plays a pivotal role in facilitating the comparison of economic activity data on global scales. Thirdly, ISIC serves as the go-to standard for categorizing data in various economic and social statistical domains, encompassing areas like national accounts, enterprise demographics, and employment. Outside the confines of traditional statistics, ISIC has been progressively incorporated in functional areas including, but not limited to, fiscal assessments and business accreditation processes. The comprehensive ISIC coding mechanism is documented within an extensive 300-page treatise, orchestrating economic activities into a hierarchically tiered, non-overlapping categorization model.

In particular, this paper leverages Large Language Models (LLMs) [7], [8], [9] for ISIC classification. Specifically, we propose a two-stage framework to deploy LLMs in our task. We first identify the most suitable model from a variety of LLMs candidates by comparing their performances in a simpler task in their pre-trained states; then we adapt the selected model to the context of ISIC classification by fine-tuning it with an additional trainable classification layer. The contributions of our work are summarized as follows:

- We study a novel problem of predicting ISIC (International Standard Industrial Classification) codes to enable waste-to-resource matching across different regions.
- We fine-tune an LLM with an additional classification layer to predict ISIC codes. We conduct extensive experiments on real-world datasets to demonstrate the effectiveness of the fine-tuned model for predicting ISIC codes.

The paper is organized as follows. Section 2 elaborates on the background of the need for ISIC classification through LLMs. Section 3 introduces the methodology framework to deploy LLMs for ISIC code prediction. Section 4 presents the experiment results and section 5 concludes the paper.

## II. BACKGROUND

This section discusses the benefits of creating a centralized knowledge repository for universal waste-to-resource matching, from where the need for ISIC classification is identified. We further introduce the ISIC framework and justify the advantages of using LLMs in our task.

### A. Waste-to-Resource Matching

Industrial symbiosis is a promising approach to achieve circular economy by reusing wastes or by-products from one company as a resource for another company [1]. Through waste-to-resource exchange, the need for virgin resources and the production of waste could be reduced, leading to both economic and environmental benefits. To facilitate waste-to-resource matching, a database comprising successful historical matches is beneficial. Some waste-to-resource databases have been developed by research society, such as MAESTRI [10] and ISDATA [11]. They usually contain information about industrial sectors (ISIC / SSIC / NACE) of waste providing and receiving companies, and details of the wastes exchanged. Unifying these historical cases not only helps analyse various aspects of industrial symbiosis, such as influences, emerging mechanisms, and driving factors, but also provides valuable insights for identifying new potential waste-to-resource matches. By referring to a successful example, a company could identify potential waste providers / receivers that a similar company (e.g., within the same industrial sector) has previously collaborated with.

Due to regional differences in the use of industrial classification standards, e.g., SSIC in Singapore and NACE in Europe, several challenges persist. First, integrating data into a unified database structure requires manual conversion among different classification standards, which is labour-intensive and time-consuming. For example, ISIC coding mechanism is extensively documented in 300 pages, making manual investigation extremely difficult. Second, mimicking historical cases also requires a company to classify its economic activities into the same standard as used in the database. Therefore, there is an urgent need to develop an automation tool to classify economic activities efficiently.

### B. ISIC Classification Framework

In this paper, we focus on ISIC as the unifying standard. Its code structure is detailed in a comprehensive 300-page codebook released by United Nations' Department of Economic and Social Affairs [4]. The ISIC framework is methodically arranged in a hierarchical manner spanning four levels: starting with 'Sections' at the apex, followed by 'Divisions' denoted by the initial two digits, 'Groups' marked by the third digit, and culminating with the 'Class', signified by the fourth digit. Table 1 below presents the hierarchical nature of ISIC with an example. Our goal is to classify any activity description into the finest ISIC code, i.e., the 4-digit ISIC code.

TABLE I
AN ISIC SAMPLE

| Level | Code | Description |
|---|---|---|
| Section | F | Construction |
| Division | 43 | Specialized Construction Activities |
| Group | 431 | Demolition And Site Preparation |
| Class | 4311 | Demolition |

### C. LLMs for domain-specific text processing

Large Language Models (LLMs), such as GPT and BERT, represent a significant advancement in Natural Language Processing (NLP). Trained on vast corpora of text data, these models can capture complex language patterns and contextual nuances. Due to the large amount of data and computing resources required to train LLMs, directly exploiting or fine-tuning pre-trained models have become a new paradigm for various domain-specific applications. For example, [12] customizes LLMs' tokenizer with equipment data and technical documents to recognize domain-specific terminologies in oil and gas industry. [13] utilizes BERT as a classification module together with an entity extraction module to extract valuable information from medical databases. To facilitate ground-truth labelling of clinical notes, [14] proposes to use LLMs together with prompt engineering and post-processing resolution. To improve classification of legal documents, [15] proposes to fine-tune DistilBERT model with domain-specific data.

In this paper, we also utilize the powerfulness of LLMs. We propose to fine-tune an LLM with domain-specific data to enable it to capture subtle differences in various activity descriptions, and hence adapt it to the context of ISIC classification. This approach enables the classification of any textual information describing economic activities into the globally recognized ISIC framework, making the creation of a unified waste-to-resource database possible.

## III. METHODOLOGY

This section elaborates on the methodology framework to deploy LLMs for ISIC classification. First, we discuss data collection method to enable the training of LLMs. Then we present the framework for LLMs development. Lastly, we discuss the evaluation method to assess the performance of the developed model.

### A. Data Collection

The EcoInvent database is a valuable resource for conducting sustainability assessments, offering a comprehensive Life Cycle Inventory (LCI) database that spans various sectors worldwide. With well-documented data for thousands of processes, products, and services, it enables users to gain a deeper understanding of the environmental impacts associated with their activities. This paper utilizes information from EcoInvent on activity names and their corresponding ISIC classifications for model training and testing. We can extract such information from the 'Undefined AO' and 'Defined AO' sections, respectively. By using the Activity Name as input and predicting its ISIC classification, we can classify any textual data describing economic activities into the unified ISIC framework.

### B. LLMs Deployment

Given the extensive diversity of ISIC codes, amounting to 182 unique categories, direct classification of these categories would result in markedly low accuracy rates. To address this problem, we designed a two-phase framework to deploy LLMs. The objective of the first phase is to identify a model suitable for ISIC classification problem among various candidate LLMs on a simpler task. The selected model is then fine-tuned with the original dataset comprising the full spectrum of 182 categories in the second phase. The whole framework is illustrated as Fig. 1.

Phase 1: model selection. In the first phase, a spectrum of advanced language models are assessed to ascertain their capabilities in the context of our classification problem. For an efficient assessment, we simplify the classification problem by utilising only the first two digits of ISIC codes out of four, effectively reducing the number of categories from 182 to a more manageable 48. The rationale behind this procedure lies in the hierarchical structure of ISIC framework. The first two digits represent the 'Division-level' classification, hence they also represent meaningful activities, providing a broad categorization of economic activities.

To preliminarily assess the models' capabilities, we leverage pre-trained LLMs and implement a cosine similarity-based approach for the classification of the 48 ISIC categories. For each candidate model, we first create a repository of embeddings of these 48 categories by feeding their textual descriptions into the model. These embeddings represent the semantic space for each ISIC category. Subsequently, for an arbitrary test sample, i.e., a pair of activity description and the corresponding ISIC code, we can obtain the embedding of the activity using the same model and compute cosine similarities between this embedding and all embeddings in the repository. The ISIC code corresponding to the highest similarity score is predicted as the category for this sample. By comparing the predicted ISIC code with the ground-truth ISIC code for all test samples, we can calculate the accuracy performance of the model. Each candidate model is evaluated using the same method, and the model with the highest accuracy is selected for the second phase.

Phase 2: Model fine-tuning. In the second phase, we proceed to fine-tune the selected model, now utilizing the original dataset comprising all 182 categories. The objective is to augment its capability to discern subtle semantic nuances and thereby optimize its classification accuracy. As shown in Fig. 1, we first freeze the selected pre-trained model and then add a new trainable classification layer on top of it. After splitting the original data into train and test sets, the model is fine-tuned using the train set and evaluated using the test set. During training process, the model's performance is continuously monitored, ensuring it is learning on the train set without overfitting to it.

### C. Evaluation Method

Our ISIC classification problem is essentially a multi-class classification problem where the categories comprise 182 unique ISIC codes. A series of commonly used metrics for multi-class classification problem are applied to evaluate the fine-tuned model. We first calculate True Positive $TP$, True Negative $TN$, False Positive $FP$, and False Negative $FN$ for each class $k \in K$, where $K$ is the set for all categories. Then we calculate the overall accuracy by dividing the total number of correct predictions over the number of all predictions, as shown in Eq. 1.

$$Accuracy = \frac{\sum_{k \in K}(TP_k + TN_k)}{\sum_{k \in K}(TP_k + TN_k + FP_k + FN_k)} \quad (1)$$

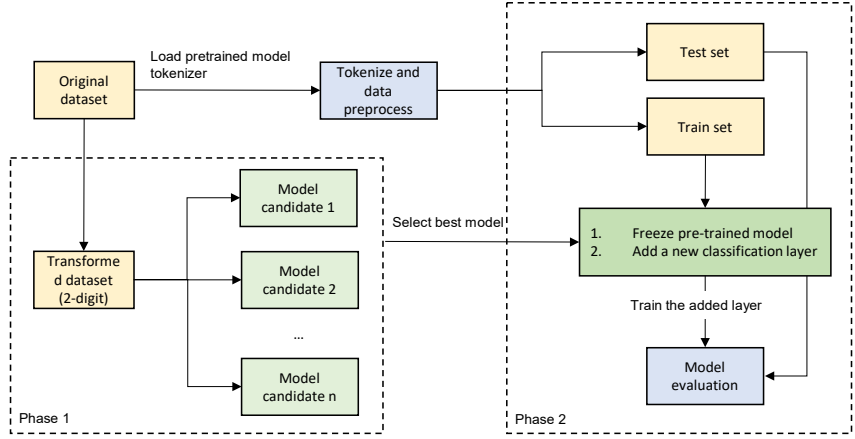

Fig. 1. Methodology framework for LLMs deployment in ISIC code classification.

We also calculate Precision, Recall and F1-score for a more comprehensive evaluation. For multi-class classification, we need to determine the averaging mechanism to aggregate performances of all classes. Micro-averaging and macro-averaging are two common methods. Micro-averaging calculates metrics globally without considering which class a sample belongs to. Macro-averaging calculates metrics for each class first and then take the average. In the context of ISIC classification where 182 categories are included, the problem of label imbalance is inevitable. Therefore, we apply a weighted macro-averaging approach to account for label imbalance, which gives higher weights to under-represented classes. To do so, we first calculate Precision, Recall and F1-score for each class $k$, as shown in Eqs. 2-4. Then we calculate the weighted averaging metrics using support $s_k$, i.e., the number of true instances for each class $k$, as shown in Eqs. 5-7. This method takes data imbalance in consideration while also reflects the prevalence of each class, making it more representative especially for imbalanced datasets.

$$Precision_k = \frac{TP_k}{TP_k + FP_k} \quad (2)$$

$$Recall_k = \frac{TP_k}{TP_k + FN_k} \quad (3)$$

$$F1_k = 2 \cdot \frac{Precision_k \cdot Recall_k}{Precision_k + Recall_k} \quad (4)$$

$$Precision_A = \frac{1}{\sum_{k \in K} s_k} \left( \sum_{k \in K} Precision_k \cdot s_k \right) \quad (5)$$

$$Recall_A = \frac{1}{\sum_{k \in K} s_k} \left( \sum_{k \in K} Recall_k \cdot s_k \right) \quad (6)$$

$$F1_A = \frac{1}{\sum_{k \in K} s_k} \left( \sum_{k \in K} F1_k \cdot s_k \right) \quad (7)$$

## IV. EXPERIMENT AND DISCUSSION

### A. Model Selection

We first present the experiments for model selection. Table 2 shows the performance of various candidate models. It is observed that the highest accuracy achieved is a mere 27.6% using GPT-2. Such suboptimal performance can be attributed to a couple of factors. First, the models, including GPT-2, were utilized in their pre-trained states without fine-tuning to our dataset or the unique nuances of the ISIC code descriptions. Pre-trained models are equipped with a generalized understanding of language. Without fine-tuning, they might not capture the domain-specific intricacies inherent in economic activity descriptions. Second, relying solely on similarities among embeddings of activity descriptions, especially in a domain as nuanced as economic activities, can lead to ambiguous matches. The semantic space of economic activities can be densely packed, where multiple activities might be semantically close but belong to different ISIC categories.

Nevertheless, the goal of the first step is to identify the most suitable model for fine-tuning. Despite the low accuracy of these models, we may still identify GPT-2 as the most promising one. A huge improvement is expected after fine-tuning the model in the second phase.

TABLE 2
COMPARITIVE ACCURACY OF LARGE LANGUAGE MODELS ON ISIC CODE PREDICTION

| Models | Accuracy |
| --- | --- |
| multi-qa-mpnet-base-cos-v1 | 18.20% |
| all-mpnet-base-v2 | 7.58% |
| paraphrase-MiniLM-L6-v2 | 21.23% |
| paraphrase-albert-small-v2 | 17.79% |
| RoBERTa | 11.35% |
| GPT-2 | 27.60% |

### B. Model Fine-tuning

Upon recognizing GPT-2 model in the first phase, the focus shifts toward the further fine-tuning of the model. By training the newly added classification layer on the train set, we fine-tune the GPT-2 model to better capture domain-specific characteristics. Specifically, we implement Cross Entropy Loss and use Adam optimizer with a learning rate of 0.001. The model was trained for 30 epochs. As shown in Fig. 2, the model has converged at the end of training. Post-refinement, the model showcased remarkable improvement. As shown in Table 3, upon evaluating the model's performance on the test set, we observe a notable accuracy of 95.28%. The similar high level of precision, recall and F1-score also indicates the effectiveness of the model. Eventually, our fine-tuned GPT-2 model is able to classify any text describing economic activities into the unified ISIC framework with high accuracy, enabling automatic classification of economic activities and facilitating waste-to-resource matching globally.

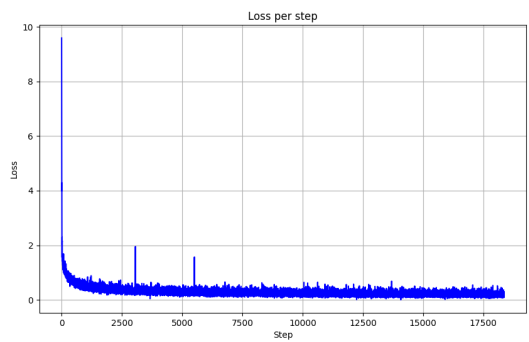

Fig. 2. Plot of loss per step during training for 50 epochs.

TABLE 3
PERFORMANCE OF THE FINE-TUNED GPT-2 MODEL ON TEST SET

| Accuracy | Precision weighted | Recall weighted | F1weighted |
|---|---|---|---|
| 95.28% | 95.37% | 95.28% | 95.27% |

## V. CONCLUSION

In this research, we embarked on exploring a novel problem of predicting the ISIC (International Standard Industrial Classification) codes for classifying textual data described economic activities, which facilitates waste-to-resource matching across different regions. A strategic adaptation of LLMs has been successfully implemented, involving fine-tuning processes augmented with an additional classification layer specifically designed for ISIC code prediction, hence leveraging the transformative capabilities of LLMs for this unique task. A cornerstone of our research methodology is the implementation of extensive experiments utilizing real-world datasets. These experiments are instrumental in unveiling the practical applicability and effectiveness of the proposed method in predicting the ISIC codes with appreciable accuracy and reliability. The outcomes of this research signify a meaningful advancement in the field, and consequently offering substantial contributions towards the enhancement of economic data analysis and classification practices.

ACKNOWLEDGMENT

This project is supported by Agency for Science, Technology and Research (A*STAR) Central Research Fund - Applied & Translational Research, and Ministry of Education, under contract RG10/23.